\documentclass{article}

\PassOptionsToPackage{numbers, compress}{natbib}

\usepackage[final]{styles/neurips_2024}

\usepackage[utf8]{inputenc} 
\usepackage[T1]{fontenc}    
\usepackage{hyperref}       
\usepackage{url}            
\usepackage{booktabs}       
\usepackage{amsfonts}       
\usepackage{nicefrac}       
\usepackage{microtype}      
\usepackage{xcolor}         

\usepackage[pdftex]{graphicx}
\usepackage{amsmath}
\usepackage{amssymb}
\usepackage{amsthm}
\usepackage[ruled, noline]{algorithm2e}
\usepackage{cleveref}

\theoremstyle{definition}
\newtheorem{definition}{Definition}[section]

\theoremstyle{remark}

\newcommand*{\R}{\mathbb{R}} 
\newcommand*{\N}{\mathbb{N}} 
\newcommand*{\X}{\mathcal{X}} 
\newcommand*{\Y}{\mathcal{Y}} 
\newcommand*{\tran}{^{\mkern-1.5mu\mathsf{T}}} 

\DeclareMathOperator*{\argmin}{arg\,min}

\title{Training Hamiltonian neural networks\\without backpropagation}
\author{%
  Atamert Rahma \\
  School of Computation, Information\\
  and Technology\\
  Technical University of Munich\\
  Munich, Germany \\
  \texttt{atamert.rahma@tum.de} \\
  \And
  Chinmay Datar \\
  School of Computation, Information\\
  and Technology\\
  Technical University of Munich\\
  Munich, Germany \\
  \texttt{chinmay.datar@tum.de} \\
  \And
  Felix Dietrich \\
  School of Computation, Information\\
  and Technology\\
  Technical University of Munich\\
  Munich, Germany \\
  \texttt{felix.dietrich@tum.de} \\
}

\author{
  Atamert Rahma\(^{1}\) \quad Chinmay Datar\(^{1,2}\) \quad Felix Dietrich\(^{1,3}\) \\
  \(^{1}\)School of Computation, Information and Technology \\
  \(^{2}\) Institute for Advanced Study \\
  \(^{3}\) Munich Data Science Institute \\
  Technical University of Munich  \\
  Munich, Germany \\
  \{atamert.rahma,chinmay.datar,felix.dietrich\}@tum.de
}

\begin{document}

\maketitle

\begin{abstract}
Neural networks that synergistically integrate data and physical laws offer great promise in modeling dynamical systems. 
However, iterative gradient-based optimization of network parameters is often computationally expensive and suffers from slow convergence. 
In this work, we present a backpropagation-free algorithm to accelerate the training of neural networks for approximating Hamiltonian systems through data-agnostic and data-driven algorithms. 
We empirically show that data-driven sampling of the network parameters outperforms data-agnostic sampling or the traditional gradient-based iterative optimization of the network parameters when approximating functions with steep gradients or wide input domains. 
We demonstrate that our approach is more than 100 times faster with CPUs than the traditionally trained Hamiltonian Neural Networks using gradient-based iterative optimization and is more than four orders of magnitude accurate in chaotic examples, including the Hénon-Heiles system.


\end{abstract}

\begin{figure}[!htb]
  \begin{center}
    \includegraphics[width=1\linewidth]{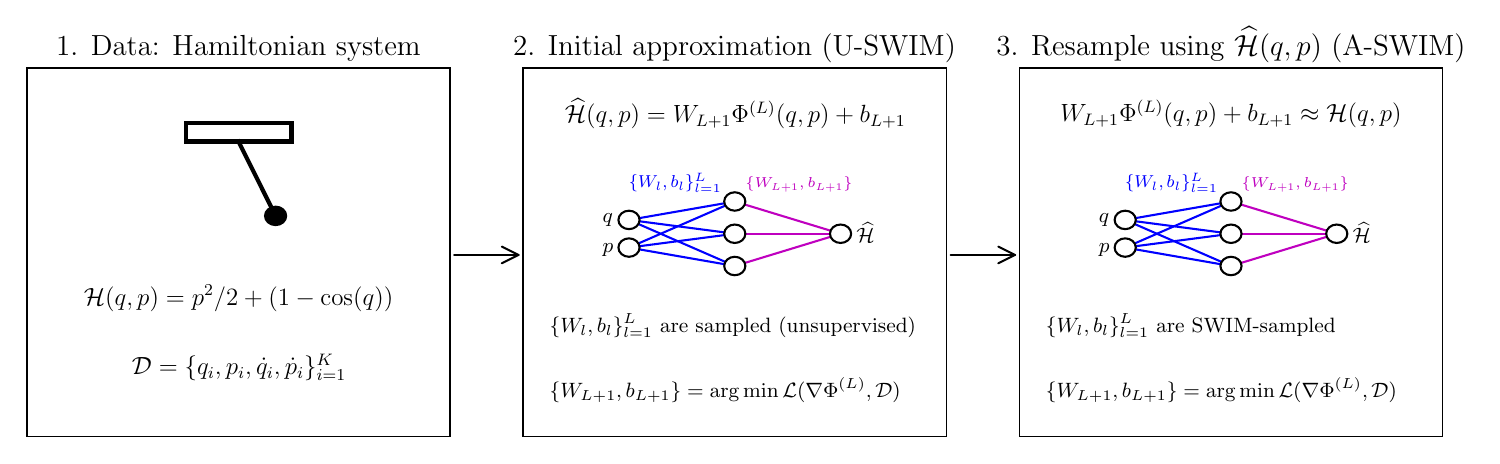}
  \end{center}
  \caption{Approximate-SWIM (A-SWIM) algorithm: This figure illustrates the process of approximating a Hamiltonian system from data, including generalized ``position'' \( q \) and ``momentum'' \( p \) coordinates, along with their time derivatives \( \dot{q} \) and \( \dot{p} \). \textbf{Left}: The given Hamiltonian system. \textbf{Center}: Sampling hidden layer weights and biases \( \{W_{l}, b_{l}\}_{l=1}^{L} \) in the unsupervised setting. \textbf{Right}: Resampling hidden layer parameters using the approximated function values \( \widehat{\mathcal{H}}(q,p) \) obtained in stage two. Note that in steps two and three, the linear layer parameters \( \{ W_{L+1}, b_{L+1} \} \) are optimized by solving a linear least-squares problem.}\label{fig:approx-swim-idea}
\end{figure}


\section{Introduction}\label{section:intro}



Learning techniques often integrate a series of inductive biases to learn inherent patterns within
data and extend the generalization capability beyond the training set.
In the context of approximating physical systems, physical
priors are crucial for capturing the system's nature, including its dynamics and underlying physical laws
\citep{watters-2017-physics-priors, de-2018-physics-priors, chang-2016-physics-priors}.
In particular, {\bf Hamiltonian Neural Networks (HNNs)}  leverage 
Hamilton's equations \citep{hamilton-1834, hamilton-1835} and re-formulation of the loss functions to learn the
conserved quantities of target systems \citep{greydanus-2019-HNN, bertalan-2019}. 
Subsequent research has introduced various extensions to HNNs, such as port-HNNs \citep{shaan-2021-port-HNN} for controlling forces and dissipative HNNs \citep{sosanya-2022-dissipative-HNN} for modeling dissipative systems.
Recent work has further extended HNNs by integrating symplectic methods~\citep{dierkes-2023,offen-2022-GP-inverse-modified-hamiltonian,ober-blobaum-2023}. Symplectic Recurrent Neural
Networks (SRNNs), proposed by \citet{chen-2019-SRNNs}, outperform traditional HNNs in long-term
trajectory prediction from time series data by incorporating symplectic integrators.
Similarly, Symplectic Ordinary Differential Equations (SymODEN) \citep{zhong-2020-symoden} and its
dissipative variant \citep{zhong-2020-dissipative-symoden} use symplectic integrators and can also
account for external forces, such as those involved in control systems.
Additionally, \citet{xiong-2022-nonseperable-symplectic-NNs} employed explicit high-order integrator schemes within an extended phase space to address non-separable Hamiltonians.

Despite the success, many challenges are becoming apparent, mainly stemming from gradient-based optimization of parameters using backpropagation. 
Backpropagation through an integrator is computationally expensive and time-consuming \citep{xiong-2022-nonseperable-symplectic-NNs}.
Moreover, iterative gradient-descent-based training approaches pose a significant challenge for
traditional neural networks due to their slow convergence rates \citep{jacobs-1988-slow-convergence}.
\citet{jakovac-2022} explored {\bf random feature models} in Hamiltonian flow approximation and
reported faster training times, although their work was limited to the data-agnostic method: Extreme
Learning Machine (ELM) \citep{huang-2004-ELM}. 
We adopt a strategy that is similar to that of \citet{bertalan-2019}, solving system identification through a linear PDE, but replacing their Gaussian Process ansatz with data-driven random features.
Our key contributions in this work are as follows:
\begin{enumerate}
    \item We explore data-agnostic \citep{huang-2004-ELM} and data-driven algorithms \citep{bolager-2023-SWIM,datar-2024a} to compute parameters of the neural networks for learning Hamiltonian functions from data \textbf{without backpropagation}.
    \item We investigate the \textbf{benefits of data-driven sampling} over data-agnostic approaches in improving accuracy for systems like single/double pendulums, Lotka-Volterra, and Hénon-Heiles.
    \item We demonstrate that data-driven sampling achieves high accuracy in the unsupervised setting \textbf{with limited data} using approximate initial function values.
\end{enumerate}

\section{Method}\label{section:method}

We consider an autonomous Hamiltonian on Euclidean space \( E=\R^{2d}\), where \( d \in \N \)
is the number of degrees of freedom of the underlying system. The Hamiltonian is defined as
\( \mathcal{H}: E \rightarrow \R \) and incorporates the laws of motion defined through Hamilton's
equations \citep{hamilton-1834, hamilton-1835}
\begin{equation}\label{eq:hamiltons-pde}
  J \cdot \nabla\mathcal{H}(q,p) - v(q,p) = \vec{0}
\end{equation}
for all \( \begin{bmatrix} q \\ p \end{bmatrix} \in E\), where \( J = \begin{bmatrix} 0_d & I_d \\ -I_d & 0_d \end{bmatrix}
\in {\{0,1\}}^{2d \times 2d} \), \(I_d \in {\{0,1\}}^{d \times d}\) is the identity matrix,
\( 0_d \) is the \(d\) by \(d\) square matrix of zeros, and \(v = {\begin{bmatrix} \dot{q} & \dot{p} \end{bmatrix}}\tran \)
is the vector field on the phase space \( E \) which only depends on generalized ``position'' \( q(t) \in \R^{d} \)
and ``momentum'' \( p(t) \in \R^{d} \) coordinates.

\paragraph{Sampling neural networks to approximate Hamiltonian functions:}
We model the Hamiltonian that we would like to approximate using a fully-connected neural network
\( \Phi \) following \Cref{def:neural-network}
with \( L \) hidden layers given data \( \mathcal{D} = \{ q_i, p_i, \dot{q}_i, \dot{p}_i \}_{i=1}^{K} \),
such that
\begin{equation}\label{eq:approximation-target}
  \widehat{\mathcal{H}}(q,p) = \Phi(q,p) = W_{L+1} \Phi^{(L)}(q,p) + b_{L+1} \overset{!}{=} \mathcal{H}(q,p) \,\, \text{for all} \,\, {\begin{bmatrix}q & p\end{bmatrix}}\tran \in E,
\end{equation}
where \( \{ W_l,b_l \}_{l=1}^{L} \) are the weights and biases of the hidden layers, \( W_{L+1} \) and
\( b_{L+1} \) are the weights and biases of the last linear layer, and
we write \( \Phi^{(l)}(\cdot) \) to represent the output of the \( l \)-th layer of the network.
To construct the hidden layers we sample the parameters \( \{W_l,b_l\}_{l=1}^{L} \) using the following data-agnostic and data-driven sampling
schemes.

\paragraph{Data-agnostic sampling:} We use the {\bf ``Extreme Learning Machine'' (ELM)} approach which is well-studied \citep{schmidt-1992-schmidt-neural-network, pao-1992-random-vector-functional-link, huang-2004-ELM, huang-2004-ELM, zhang-2012-augmented-ELM, leung-2019-ELM, jakovac-2022} along with the error bounds and approximation capabilities
\citep{huang-2006-universal-approx-ELM, rahimi-2008-random-feature, zhang-2012-augmented-ELM, leung-2019-ELM}. 
In this approach, we sample the weights \( \{W_{l}\}_{l=1}^{L} \) from the standard normal distribution and biases \( \{ b_{l} \}_{l=1}^{L} \) from the uniform distribution for all hidden layers. 

\paragraph{Data-driven sampling:} We use the \textbf{``Sample Where It Matters'' (SWIM)} algorithm by \citet{bolager-2023-SWIM}. Here, each weight and bias pair in the hidden layers is constructed using a data point pair sampled from the input space so that 
\(w_{l,i} = s_1 \frac{x_{l-1, i}^{(2)} - x_{l-1, i}^{(1)}}{ {\lVert x_{l-1,i}^{(2)} - x_{l-1,i}^{(1)} \rVert}^2 }\) and \(b_{l,i} = - \langle w_{l,i}, x_{l-1,i}^{(1)} \rangle - s_2 \), 
where \( (w_{l,i}, b_{l,i}) \) are the \( i \)-th row of the parameters in the \( l \)-th hidden layer,
\( (x_{l-1,i}^{(1)}, x_{l-1,i}^{(2)}) \) are the outputs of the hidden layer \( (l-1) \) of the data point pair
sampled for constructing the parameters \( (w_{l,i}, b_{l,i}) \), and \( (s_1, s_2) \) are constants depending on
the activation function used in the associated hidden layer. SWIM-sampled neural networks are discussed in more detail in \Cref{def:swim-sampled-nn}.

The construction associates each hidden layer parameter with an input pair sampled from the given data points from the input space \(\X\). 
In the unsupervised setting, we sample the pairs of points uniformly randomly and refer to this as {\bf Uniform-SWIM (U-SWIM)}. 
In the supervised setting, we sample the points with a density proportional to the finite differences 
\( {\lVert \mathcal{H}(x_0^{(2)}) - \mathcal{H}(x_0^{(1)}) \rVert}/{\lVert x_{l-1}^{(2)} - x_{l-1}^{(1)} \rVert} \) and refer to this as {\bf SWIM} \citep{bolager-2023-SWIM}. 
The intuition behind the density in SWIM is to place more basis functions in the part of the space domain where the gradient of the underlying function is large. 
To retain this efficient placement of basis functions using SWIM, but in the unsupervised setting, we propose an adaptive approach in which we first compute an initial approximation and then use the SWIM algorithm to re-sample the network parameters as illustrated in \Cref{fig:approx-swim-idea}. We refer to this method as {\bf Approximate-SWIM (A-SWIM)} and use U-SWIM for the initial approximation.


After sampling all the hidden layer parameters, we discuss finding the optimal parameters for the last
linear layer of the network. Differentiating \Cref{eq:approximation-target} with respect
to input \(x\) results in \(W_{L+1} \nabla \Phi^{(L)}(x) \overset{!}{=} \nabla \mathcal{H}(x) \overset{\text{{\tiny \cref{eq:hamiltons-pde}}}}{=}
J^{-1} \dot{x}\) 
for a single data point \(x = {\begin{bmatrix} q & p \end{bmatrix}}\tran \in E \). Similar to \citep{bertalan-2019}, we set up a fully linear system using the given data, where we replace the basis functions of the approximation with the outputs of the last hidden layer of the network \( \Phi \):
\begin{equation}\label{eq:fully-linear-system}
  \underbrace{
    \begin{bmatrix}
      \nabla \Phi^{(L)}(x_1)& \cdots  & \nabla \Phi^{(L)}(x_K) & \Phi^{(L)}(x_0) \\
      0 & \cdots & 0 & 1\\
    \end{bmatrix}\tran
  }_{A \in \R^{(2dK+1) \times (N_L+1)}}
  \cdot
  \underbrace{
    \begin{bmatrix}
      W_{L+1}\tran \\
      b_{L+1}
    \end{bmatrix}
  }_{w \in \R^{N_L+1}}
  \overset{!}{=}
  \underbrace{
    \begin{bmatrix}
      J^{-1} \dot{x}_1  &
      \cdots & J^{-1} \dot{x}_K & \mathcal{H}(x_0)
    \end{bmatrix}\tran,
  }_{u \in \R^{2dK+1}}
\end{equation}

where we write \( N_{l} \) for the width of the \( l \)-th hidden layer. We note that we assume we
know the true Hamiltonian value \( \mathcal{H}(x_0) \) for a single data point \(x_0 \in E\) to
fix the integration constant \( b_{L+1} \) (which is only a scalar). We also assume we know the true time derivatives \(\dot{x}\) in this work to avoid discretization errors. We shortly discuss error correction techniques in \Cref{section:appendix:numerical:error-correction} if finite differences are used to retrieve \(\dot{x}\), e.g. from trajectory data.
\Cref{eq:fully-linear-system} gives rise to a well-studied convex optimization problem
which can be solved by using the linear least squares
\( \begin{bmatrix} W_{L+1}\tran & b_{L+1} \end{bmatrix}\tran =\argmin_w { \lVert A w - u \rVert }^2.\)

The details of the algorithms for constructing sampled HNNs using the samplers ELM, U-SWIM, or A-SWIM are provided
in \Cref{alg:sampled-hnn}.
The gradient \( \nabla \Phi^{(L)} \) is computed analytically as we use differentiable activation functions (\(\sigma = \tanh \)).

\section{Numerical results and discussion}\label{section:results-and-discussion}

In this section, we report our numerical results, including approximations of different Hamiltonian systems with
different configurations. Details regarding the experiment setup, used Hamiltonian target functions,
software versions, and the evaluated error function are explained in \Cref{appendix:numerical-experiments}. The code is available open-source at \url{https://github.com/AlphaGergedan/Sampling-HNNs}.

\begin{figure}[!htb]
  \begin{center}
    \includegraphics[width=1\linewidth]{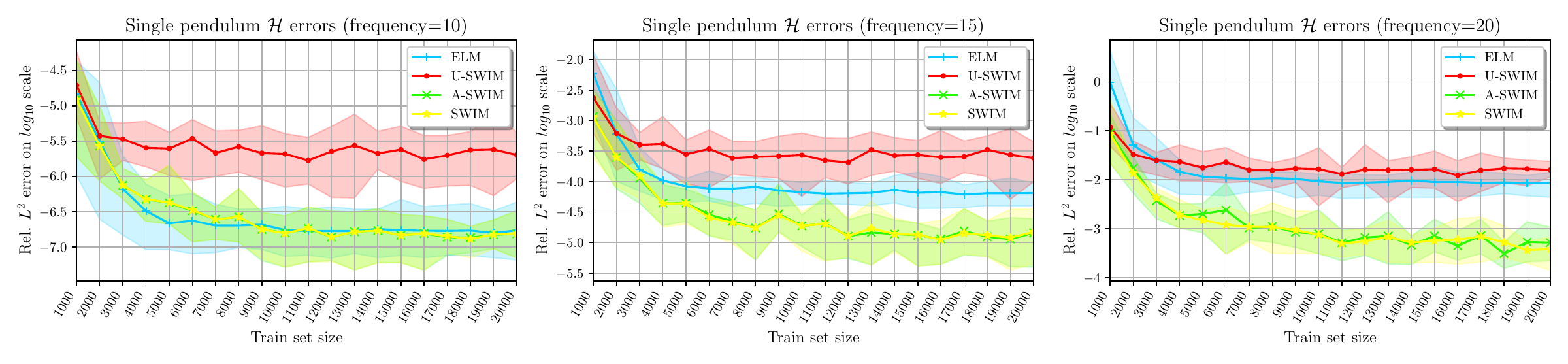}
  \end{center}
  \caption{Single pendulum (with frequency parameter) approximation errors are plotted.}\label{fig:single-pendulum-freq-errors}
\end{figure}

\paragraph{Single pendulum and Lotka-Volterra experiments:}
\Cref{table:single-pendulum-lotka-volterra-experiments-summary} summarizes the experiments for
the single pendulum and Lotka-Volterra \citep{fernandes-1995-lotka-volterra} Hamiltonian functions. The network width is scaled for each experiment in the table, and \Cref{fig:single-pendulum} and \Cref{fig:lotka-volterra-errors} describe how the error decays with the network width. 
Firstly, we observe that the losses obtained with gradient-free training are 3-9 orders of magnitude better. 
Secondly, we observe that as the domain size or the frequency of the single pendulum increases, SWIM is usually more accurate and also needs fewer neurons to attain a certain accuracy compared to ELM as it exploits the data to place basis functions efficiently (more where the solution gradient is large).  
Thirdly, the accuracies obtained with A-SWIM are much better than those with U-SWIM and are very close to those obtained with SWIM, even though the true values of the Hamiltonian are not available in A-SWIM. 
Lastly, \Cref{fig:single-pendulum-freq-errors} demonstrates that increasing the frequency parameter in the single-pendulum experiment leads to steep solution gradients, where A-SWIM and SWIM prove to be far more accurate than ELM.


\begin{table}[!htb]
  \caption{Single pendulum and Lotka-Volterra approximations are summarized. In all the listed experiments,
  A-SWIM and SWIM errors have the same values up to the order that we list here therefore we list them
  together in the column (A-)SWIM.
  Network width is set to \(1000\), domain and other model parameters are set according to
  \Cref{table:domain-params} and \Cref{table:approx-params}, respectively.}\label{table:single-pendulum-lotka-volterra-experiments-summary}
  \centering
  \begin{tabular}{llllll}
    \toprule
    System name                     & Hamiltonian                                                                                             & Domain                              & HNN               &   ELM            &   (A-)SWIM          \\
    \midrule
    Single pendulum                 & \cref{eq:single-pendulum}                                                                       & \([-2\pi, 2\pi]\times[-1,1]\)       & 2.17E-03 &   \textbf{1.49E-11}          &   3.62E-10          \\ 
    Single pendulum                 & \cref{eq:single-pendulum}                                                                       & \([-2\pi, 2\pi]\times[-6,6]\)       & 7.37E-04           &   9.82E-08 &   \textbf{1.55E-09}          \\ 
    Lotka-Volterra                  & \cref{eq:lotka-volterra-zero-centered}                                                          & \([-2,2]\times[-2,2]\)              & 2.35E-03 &   \textbf{3.04E-12}          &   1.48E-10          \\ 
    Lotka-Volterra                  & \cref{eq:lotka-volterra-zero-centered}                                                          & \([-5,5]\times[-5,5]\)              & 1.38E-03          &   1.02E-08          &   \textbf{7.99E-09} \\ 
    Lotka-Volterra                  & \cref{eq:lotka-volterra-five-centered}                                                          & \([0,8]\times[0,8]\)                & 2.63E-03          &   3.27E-06          &   \textbf{1.51E-08} \\ 
    \bottomrule
  \end{tabular}
\end{table}

\paragraph{Chaotic systems:}
We summarize our experiments with the double pendulum and Hénon-Heiles \citep{henon-1964-henon-heiles} systems in \Cref{table:chaotic-system-approx}. 
The key observation is that the gradient-free training with A-SWIM is more than 100 times faster (with CPUs) compared to the gradient-based optimization of HNNs for the same error in the double pendulum experiment and four orders of magnitude lower error in the Hénon-Heiles experiment.
Moreover, if the initial approximation with (U-SWIM) is sufficiently accurate, A-SWIM can closely match the SWIM method's performance. We emphasize that while SWIM uses true Hamiltonian values for sampling, A-SWIM relies solely on approximate values for this process, which results in approximately double the training time.
\begin{table}[!htb]
  \caption{Summary of results for chaotic system approximations. Note that we report the CPU time. Please refer to \Cref{table:chaotic-system-approx-params} and \Cref{table:domain-params} for details on models.}\label{table:chaotic-system-approx}
  \centering
  \begin{tabular}{lll|ll}
    \toprule
              &     \multicolumn{2}{c}{Double pendulum}             &   \multicolumn{2}{c}{Hénon-Heiles}                  \\
    \midrule
    Method    &     Training time (s)   &   Rel. \(L^2\) error  &   Training time (s)   &   Rel. \(L^2\) error            \\
    \midrule
    HNN       &     10485.4              &   \textbf{3.62E-03}                &   13140.6              &   6.68E-04                  \\
    ELM       &     43.1                &   5.69E-03                &   46.0                &   2.07E-02                  \\
    U-SWIM    &     \textbf{39.6}                &   5.00E-03                &   \textbf{41.7}                &   8.22E-08                  \\
    A-SWIM    &     85.2       &   4.08E-03       &   89.4       &   \textbf{6.80E-08}         \\
    SWIM      &     40.0                &   4.18E-03                &   41.9                &   \textbf{6.80E-08}         \\
    \bottomrule
  \end{tabular}
\end{table}

\paragraph{Conclusion and future work:}
We presented a framework for approximating Hamiltonian functions using sampled neural networks without requiring iterative parameter optimization via backpropagation. 
Our approach (A-SWIM) is more than two orders of magnitude faster to train on CPUs than gradient-based optimization of HNNs in the chaotic systems we consider, and it achieves more than four orders of magnitude greater accuracy in most examples.
Our evaluation demonstrates that data-driven sampling via SWIM outperforms data-agnostic methods when the function being approximated has steep gradients or large input domains.
We note that our approach requires solving a large linear system. In higher dimensional examples that require a lot of computational requirements, one can rely on HPC resources and iterative solvers.
%
%
In the future, we intend to extend this work to dissipative systems, as this work assumes symplecticity, while most real-world systems are dissipative. 
Another important direction is to extend our algorithm to handle noisy data.
Lastly, the longer training times reported in \citep{xiong-2022-nonseperable-symplectic-NNs} may be compensated by utilizing sampled networks to learn the flow map of Hamiltonian systems where using backpropagation is computationally expensive.

\begin{ack}



The authors gratefully acknowledge the computational and data resources provided by the Leibniz Supercomputing Centre (www.lrz.de). 
We also thank the anonymous reviewers at NeurIPS for their constructive feedback.
F.D. acknowledges funding by the German Research Foundation---project 468830823, and association to DFG-SPP-229. C.D. is partially funded by the Institute for Advanced Study (IAS) at the Technical University of Munich.
\end{ack}


{
\small
\bibliographystyle{plainnat}
\bibliography{lib}
}
\clearpage

%
%

%
%
%
%
%
%


\setcounter{page}{1}

\appendix

\renewcommand{\thefigure}{\thesection.\arabic{figure}}
\renewcommand{\theequation}{\thesection.\arabic{equation}}
\renewcommand{\thealgocf}{\thesection.\arabic{algocf}}
\renewcommand{\thetable}{\thesection.\arabic{table}}

\section*{Appendix}

\section{Mathematical framework}\label{section:appendix:math}

\paragraph{Feed-forward neural networks:}
In this paper, we work with feed-forward neural networks configured for regression, i.e., no activation
is used in the output layer to approximate a Hamiltonian. We define the notation used for neural
networks in this work in \Cref{def:neural-network}.

\begin{definition}\label{def:neural-network}
  Let \(\X \subseteq \R^{D}\) be an input space, and \(\Y \subseteq \R\) a one-dimensional output space.
  We write
  \begin{equation*}
    \Phi^{(l)}(x) = \begin{cases}
      x,                                          & \text{for } l=0 \\
      \sigma(W_{l} \Phi^{(l-1)}(x) + b_l),        & \text{for } 0 < l \leq L \\
      W_{L+1} \Phi^{(L)}(x) + b_{L+1},            & \text{for } l=L+1
    \end{cases}
  \end{equation*}
  as the output of the \(l\)-th layer of a network \(\Phi\) with \(L\) hidden layers, where
  \begin{itemize}
    \item \(\sigma: \R \rightarrow \R\) is an activation function applied element-wise,
    \item \(\{W_l, b_l\}_{l=1}^{L+1}\) are the parameters of \(\Phi\): weights and biases, where
      \(W_l \in \R^{N_l \times N_{l-1}}\) and \(b_l \in \R^{N_l}\). \(N_l\) is the number of neurons
      in the \(l\)-th layer with \(N_0 = D\) and \(N_{L+1} = 1\).
  \end{itemize}
  We write
  \(\Phi(x) \in \Y\) for the network output given a single data point \(x \in \X\) and \(\Phi(X) \in \Y^{K}\) given
  an input matrix \(X \in \X^{K \times D}\).
\end{definition}

\paragraph{``Sample Where It Matters'' (SWIM):}
Our work is closely related to \citep{bolager-2023-SWIM} as we use their SWIM algorithm for the data-driven sampling of the hidden layer parameters. In \Cref{def:swim-sampled-nn}
we summarize a SWIM-sampled network from \citep{bolager-2023-SWIM} to provide the overall idea of
using data to sample the hidden layer parameters.

\begin{definition}\label{def:swim-sampled-nn}
  Let \( \Phi \) be a network as defined in \Cref{def:neural-network}. For \( l=1, \dots, L \)
  let \( x_{0,i}^{(1)}, x_{0,i}^{(2)} \) be pairs of points sampled over \(\X \times \X\). If the weights
  and biases of each layer \(l = 1, 2, \dots, L\) and neuron \(i = 1, 2, \dots, N_l\) have the form
  \begin{equation*}
    w_{l,i} = s_1 \frac{x_{l-1,i}^{(2)} - x_{l-1,i}^{(1)}}{ { \lVert x_{l-1,i}^{(2)} - x_{l-1,i}^{(1)} \rVert }^2}
    \hspace{0.25em}, \hspace{2em}
    b_{l,i} = - \langle w_{l,i}, x_{l-1,i}^{(1)} \rangle - s_2 \hspace{0.25em},
  \end{equation*}
  then we say the hidden layer parameters \(\{ W_l, b_l \}_{l=1}^{L}\) are SWIM-sampled and \( \Phi \) is a
  SWIM-sampled network. \(\lVert \cdot \rVert\) is the \(L^2\) norm, \(\langle \cdot, \cdot \rangle\)
  the inner product, and
  \begin{itemize}
    \item \( s_1, s_2 \in \R \) are constants to place the outputs of the activation function for every
          input pair \(\{ x^{(1)}, x^{(2)} \}\). For \( \tanh \), the only activation function that we use
          in our numerical experiments, \( s_1 = 2 s_2 \) and \( s_2 = \nicefrac{\ln(3)}{2} \) are set,
          which implies \( \sigma(x^{(1)}) = \nicefrac{1}{2} \) and \( \sigma(x^{(2)}) = \nicefrac{-1}{2} \),
          and \( \sigma(\nicefrac{ (x^{(1)} + x^{(2)}) }{2} ) = 0\),
    \item \( x^{(k)}_{l-1, i} = \Phi^{(l-1)}(x^{(k)}_{0, i}) \) for \( k \in \{ 1, 2 \} \) and
          \( x^{(1)}_{l-1, i} \neq x^{(2)}_{l-1, i} \),
    \item we write \( w_{l,i} \) for the \(i\)-th row of \(W_l\) and \(b_{l,i}\) for the \(i\)-th entry of \(b_l\).
  \end{itemize}
\end{definition}
The authors also provide a probability distribution for the data points in the input space \(\X\)
(see~\Cref{def:prob-dist}) to sample where it matters: at large gradients. The supervised SWIM
algorithm we evaluate utilizes this distribution.

\begin{definition}\label{def:prob-dist}
  Let \(\X\) be an input space and let \(\Y = f(\X)\) be the function space given a function \(f: \X \rightarrow \Y\).
  When sampling a SWIM-sampled network as defined in \Cref{def:swim-sampled-nn}, for hidden
  layers \(l = 1, \dots, L\) the probability density \(p^{\epsilon}_{l}\) to sample pairs of points
  to be used for the corresponding hidden layer can be defined through the proportionality
  \begin{equation*}
    p_l^{\epsilon} \bigg( x_0^{(1)}, x_0^{(2)} | \{W_j, b_j\}_{j=1}^{l-1} \bigg) \propto \begin{cases}
      \frac{ { \lVert f(x_0^{(2)}) - f(x_0^{(1)}) \rVert }_{\Y}}{\max\Big\{ {\lVert x_{l-1}^{(2)} - x_{l-1}^{(1)} \rVert}_{\X_{l-1}}, \,\, \epsilon \Big\}},   & \text{for } x_{l-1}^{(1)} \neq x_{l-1}^{(2)} \\
      0,                                                                                      & \text{otherwise}
    \end{cases},
  \end{equation*}
  where
  \begin{itemize}
    \item \(x_0^{(k)} \in \X\) and \(x_{l-1}^{(k)} = \Phi^{(l-1)}(x_0^{(k)})\) for \(k \in \{1,2\}\)
          with the sub-network \(\Phi^{(l-1)}\) parametrized by SWIM-sampled parameters \(\{W_j, b_j\}_{j=1}^{l-1}\) which
          can be defined recursively using the density function \(p_{l-1}^{\epsilon}\) for \(L > 1\),
    \item \(\X_{l-1} = \Phi^{(l-1)}(\X)\) is the image after \((l-1)\) layers,
    \item assuming distinct inputs to the network we have \(\epsilon = 0\) for \(l=1\) and \(\epsilon > 0\) otherwise.
          We only experiment with shallow networks in this paper so we do not have to set this parameter
          in our experiments,
    \item the norms \({\lVert \cdot \rVert}_{\Y}\) and \({\lVert \cdot \rVert}_{\X_{l-1}}\) are arbitrary
          over their respective space, we choose the \(L^{\infty}\) norm for \({\lVert \cdot \rVert}_{\Y}\)
          and the \(L^2\) norm for \({\lVert \cdot \rVert}_{\X_{l-1}}\) in our experiments.
  \end{itemize}
  The probability density above can only be used in a supervised setting. The networks that we
  construct that use this distribution using the true function values \(f(\cdot)\)
  are denoted as {\bf SWIM}, and the ones that use this distribution using approximate function
  values are denoted as {\bf A-SWIM}. For the initial approximation, we use the uniform distribution
  over the pairs of points in the input space \(\X\). The networks that are
  constructed using the uniform distribution are denoted as {\bf U-SWIM}.
\end{definition}

\begin{algorithm}[!htb]
  \caption{Our proposed HNN sampling algorithm is illustrated. \(\mathcal{L}\) is a loss function, which
  in our case is the \(L^2\) loss, and \(\argmin \mathcal{L}(\cdot, \cdot)\) becomes a linear optimization problem
  that we solve using the least squares solution \( \argmin_w { \lVert A w - u \rVert }^2\) of the linear system in \Cref{eq:fully-linear-system}.}\label{alg:sampled-hnn}
  \vspace{0.5em}
  \KwData{
    \( \{ x_i, \dot{x}_i \}_{i=1}^{K}\,,\,\,\{ x_0, \mathcal{H}(x_0) \}\,,\,\,\emph{sampler} \in \{ \texttt{ELM}, \texttt{U-SWIM}, \texttt{A-SWIM} \}\) \\
  }
  \vspace{0.5em}
  \( \Phi^{(0)}(X) = X \)\;
  \For{\( l = 1,2,\dots,L \)}{
    \( W_l \in \mathbb{R}^{N_l \times N_{l-1}}, b_{l} \in \mathbb{R}^{N_{l}} \)\;
    \eIf{sampler \emph{is \texttt{ELM}}}{
      \( W_l, b_l \gets \texttt{ELM}() \)\;
    }
    {
      \( W_l, b_l \gets \texttt{U-SWIM}(\Phi^{(l-1)}(X)) \)\;
      \(\Phi^{(l)}(\cdot) \gets \sigma(W_l \Phi^{(l-1)}(\cdot) + b_l) \)\;
    }
  }
  \( \text{compute } \nabla\Phi^{(L)} \)\;
  \( W_{L+1}, b_{L+1} \gets \argmin \mathcal{L}(\nabla \Phi^{(L)}, \mathcal{D}) \)\;
  \If{sampler \emph{is \texttt{A-SWIM}}}{
    \( \widehat{\mathcal{H}}(X) \gets \Phi(X) \)\;
    \For{\( l = 1,2,\dots,L \)}{
      \( W_l \in \mathbb{R}^{N_l \times N_{l-1}}, b_{l} \in \mathbb{R}^{N_{l}} \)\;
      \( W_l, b_l \gets \texttt{SWIM}(\Phi^{(l-1)}(X), \widehat{\mathcal{H}}(X)) \)\;
      \(\Phi^{(l)}(\cdot) \gets \sigma(W_l \Phi^{(l-1)}(\cdot) + b_l) \)\;
    }

    \( \text{compute } \nabla\Phi^{(L)} \)\;
    \( W_{L+1}, b_{L+1} \gets \argmin \mathcal{L}(\nabla \Phi^{(L)}, \mathcal{D}) \)\;
  }
  \Return \( \Phi \)\;
\end{algorithm}

\section{Numerical experiments}\label{appendix:numerical-experiments}

The experiments in this section are conducted on the Linux cluster segment CoolMUC-3 at Leibniz
Supercomputing Centre (LRZ). CoolMUC-\( 3 \) provides \( 148 \) nodes (\( 64 \) cores per node) running
at the nominal frequency of \( 1.3 \) GHz and with \( \approx 96 \) GB memory (bandwidth \( 80 \) GB/s ).
Note that we have only used a single node with \( 64 \) threads in all the experiments.

For a fair evaluation, we train the networks and take the mean error values of \(10\) randomly conducted
runs, which include different seeds for the model and the generated distinct train and test sets.
We use the linear solver \verb+numpy.linalg.lstsq+ for solving the least-squares optimization problem
with python and numpy \citep{harris-2020-numpy} versions \(3.12.4\) and \(1.26.4\), respectively. We train the HNNs in our experiments with the HNN loss \citep{greydanus-2019-HNN} using the Adam optimizer implementation \verb+torch.optim.Adam+ from torch \citep{pytorch} version \(2.4.0\). The weights of the HNNs are initialized using the Xavier normal distribution \citep{glorot-2010-xavier-weight-init}, and the biases of the HNNs are zero-initialized. Similarly to the sampled HNNs, we set the integration constant, i.e., the last layer bias, accordingly assuming we know the true function value for a single data point. We used double precision (float64, both in numpy and pytorch) in our experiments. For more information regarding other software versions, random seeds, and the implementation, please refer to the open-source code repository available at \url{https://github.com/AlphaGergedan/Sampling-HNNs}.

In all the figures, we plot the mean error values of multiple runs and paint the area around this error curve using the minimum and maximum error values to demonstrate the consistency of the methods using the Matplotlib library \citep{matplotlib}.

In all the experiments, we used SWIM sampling with the approximate values resulting from U-SWIM for the A-SWIM method. For comparison, we also used SWIM sampling with the true function values and denoted this method as SWIM in the figures.

\paragraph{Error function:}

We compare the relative (rel.) \( L^2 \) error
\begin{equation*}
  \sqrt{\frac{\sum_i{ { ( \mathcal{H}(q_i,p_i) - \widehat{\mathcal{H}}(q_i,p_i) ) }^2 }}{\sum_i{ {\mathcal{H}(q_i,p_i)} }^2 }}
\end{equation*}
when analyzing and evaluating the approximations using all the points in the test set, where \(\widehat{\mathcal{H}}\)
represents the final approximation (trained model) and \(\mathcal{H}\) the true Hamiltonian. In all
tables, the mean error value of \(10\) randomly conducted runs is given for each entry for sampled networks.

\begin{figure}[!htb]
  \begin{center}
    \includegraphics[width=1\linewidth]{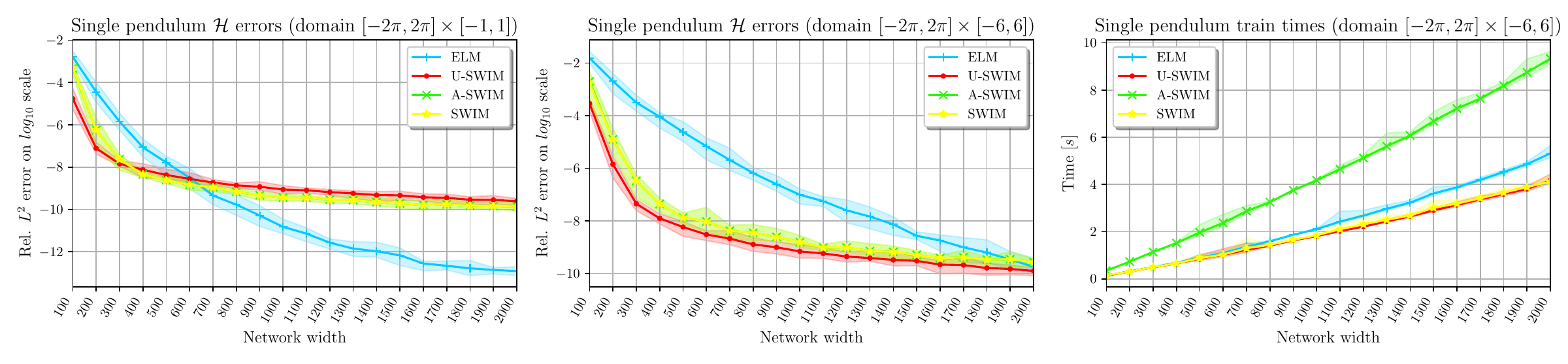}
  \end{center}
  \caption{Single pendulum approximation errors and training times for the larger domain are plotted. See \Cref{table:domain-params} and \Cref{table:approx-params} for domain and model parameters.}\label{fig:single-pendulum}
\end{figure}

\begin{figure}[!htb]
  \begin{center}
    \includegraphics[width=1\linewidth]{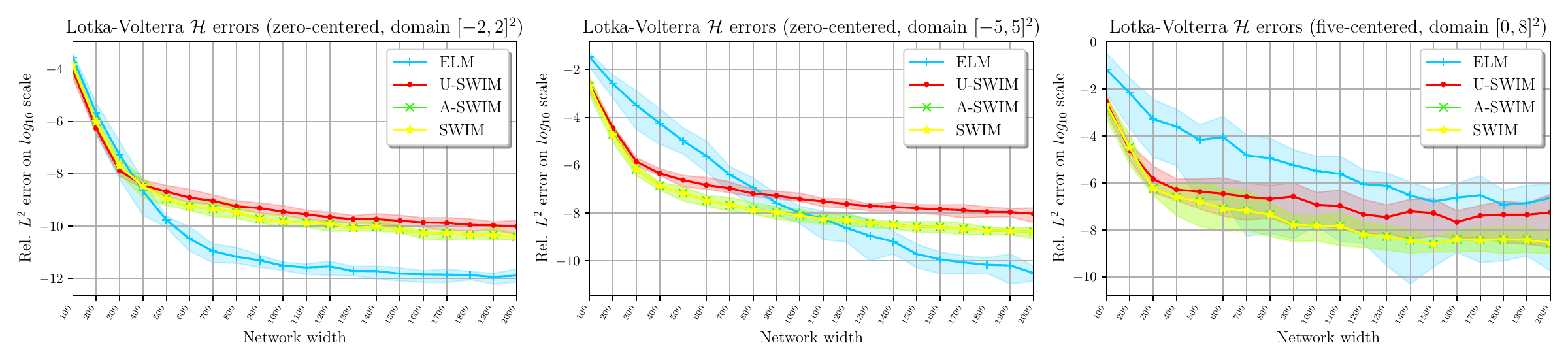}
  \end{center}
  \caption{Lotka-Volterra approximation errors are plotted. The target Hamiltonian in the left and center plots
  has an equilibrium near the zero-vector, whereas the target Hamiltonian in the right plot has an equilibrium
  around five in both dimensions. Domain and model parameters are set according to \Cref{table:domain-params} and \Cref{table:approx-params}}\label{fig:lotka-volterra-errors}
\end{figure}

\paragraph{Target Hamiltonian functions:}
Target Hamiltonian functions, which we aim to approximate in this work, are the following. The single pendulum Hamiltonian
\begin{align}
  \mathcal{H}(q,p) &= \frac{p^2}{2ml^2} + mgl (1 - \cos(q)) \nonumber \\
                   &= \frac{p^2}{2} + (1 - \cos(q)) \label{eq:single-pendulum}
\end{align}
for the experiments in \Cref{fig:single-pendulum}, where we set the constants
(mass \(m\), link length \(l\), gravitational acceleration \(g\)) to \(1\).
The single pendulum is extended with a frequency parameter resulting in the function
\begin{equation}\label{eq:single-pendulum-with-frequency}
  \mathcal{H}(q,p) = \frac{p^2}{2} + (1 - \cos(fq))
\end{equation}
for the experiments in \Cref{fig:single-pendulum-freq-errors}. The Lotka-Volterra \citep{fernandes-1995-lotka-volterra} Hamiltonian
\begin{equation*}
  \mathcal{H}(q,p) = \beta e^q - \alpha q + \delta e^p - \gamma p
\end{equation*}
with parameters
\(\beta=-1, \alpha=-2, \delta=-1, \gamma=-1\) and
\(\beta=0.025, \alpha=3.5, \delta=0.07, \gamma=10\)
for the zero-centered and five-centered domains, respectively, resulting in the functions
\begin{equation}\label{eq:lotka-volterra-zero-centered}
  \mathcal{H}(q,p) = - e^q + 2 q - e^p + p
\end{equation}
and
\begin{equation}\label{eq:lotka-volterra-five-centered}
  \mathcal{H}(q,p) = 0.025 e^q - 3.5 q + 0.07 e^p - 10 p
\end{equation}
for the experiments in \Cref{fig:lotka-volterra-errors}. The double pendulum Hamiltonian
\begin{align}
  \mathcal{H}(q,p) &= \frac{m_2 l_2^2 p_1^2 + (m_1 + m_2)l_1^2 p_2^2 - 2m_2 l_1 l_2 p_1 p_2 \cos(q_1 - q_2)}{2 m_2 l_1^2 l_2^2 (m_1 + m_2 \sin^{2}(q_1 - q_2))} \nonumber \\
                   &- (m_1 + m_2) g l_1 \cos(q_1) - m_2 g l_2 \cos(q_2) \nonumber \\
                   &= \frac{p_1^2 + 2p_2^2 - 2p_1p_2 \cos(q_1-q_2)}{2(1+\sin^{2}(q_1-q_2))} - 2 \cos(q_1) - \cos(q_2), \label{eq:double-pendulum}
\end{align}
where we set the constants
(mass \(m_1\) and \(m_2\), link lengths \(l_1\) and \(l_2\), gravitational acceleration \(g\)) to \(1\) for the experiments in \Cref{table:chaotic-system-approx}.
The Hénon-Heiles \citep{henon-1964-henon-heiles} Hamiltonian
\begin{equation}\label{eq:henon-heiles}
  \mathcal{H}(q,p) = \frac{1}{2} (p_1^2 + p_2^2) + \frac{1}{2} (q_1^2 + q_2^2) + \alpha (q_1^2 q_2 - \frac{1}{3}q_2^3),
\end{equation}
where we set the bifurcation parameter
\(\alpha=1\) for the experiments in \Cref{table:chaotic-system-approx}.

\paragraph{Single pendulum:}
We show that all the methods can reach very low approximation errors in \Cref{fig:single-pendulum}
when approximating the single pendulum Hamiltonian.

Especially in the larger domain and with smaller network widths, SWIM sampling outperformed ELM.
On the other hand, while the SWIM methods' performance remained consistent,
ELM could reach lower approximation errors in both domains with large network widths
(more than \(600\) in the smaller, \(2000\) in the larger domain). In the larger domain, the gradients
of the system move away from zero, and with large gradients where the function values change quickly
in the target function, the SWIM methods might have constructed better (more accurate and consistent) weights using
the data-driven scheme with small network widths. Also, ELM may need larger network widths to be able to
cover the input space uniformly in larger domains. Remarkably, A-SWIM can match the SWIM method's
accuracy in all the settings with an additional training time cost of resampling after the initial
approximation. Again, we emphasize that A-SWIM does not have access to the true function values but uses
approximate function values to utilize the SWIM sampling scheme, whereas SWIM requires the true
function values.

\paragraph{Lotka-Volterra:}
We experiment with Lotka-Volterra Hamiltonians in different settings in \Cref{fig:lotka-volterra-errors}.
Similar to the single pendulum experiments, the ELM method performed better with large network widths than the
SWIM sampling, however, its performance has declined in the larger domains where the gradients are larger.
Also, note that when the equilibrium changes to a vector close to 5 in both dimensions,
ELM method becomes less consistent.

\begin{figure}[!htb]
  \begin{center}
    \includegraphics[width=1\linewidth]{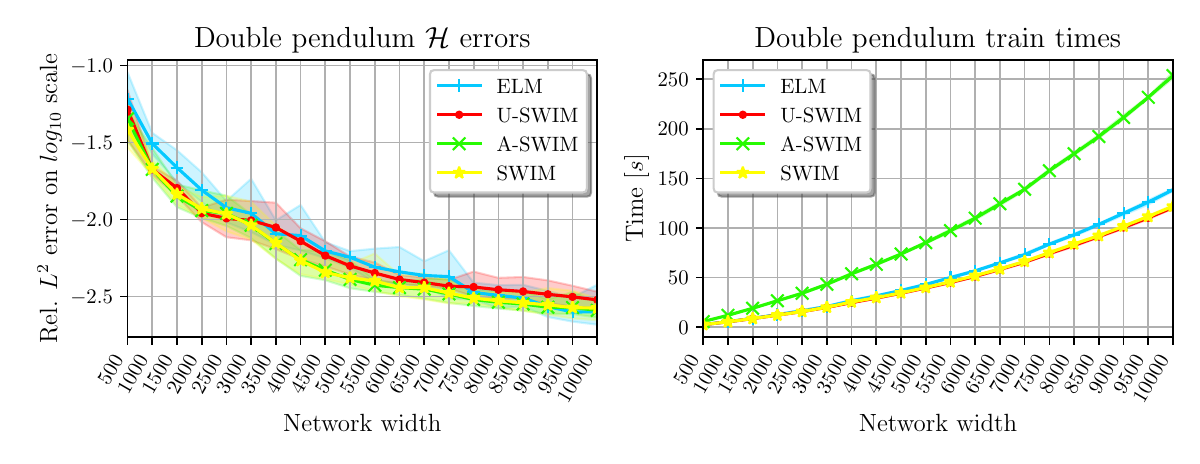}
  \end{center}
  \caption{Double pendulum approximation errors and training times are displayed. Network width was
  scaled and other model parameters were set as described in \Cref{table:chaotic-system-approx-params}. Domain information is listed in \Cref{table:domain-params}.}\label{fig:double-pendulum}
\end{figure}

\paragraph{Chaotic-systems:}\label{appendix-b:chaotic-systems}
In addition to the results in \Cref{table:chaotic-system-approx}, we provide network scaling for
the double pendulum and note that all the methods perform similarly in \Cref{fig:double-pendulum}. To efficiently train HNNs using backpropagation, we have also tried using the NVIDIA GeForce RTX 3060 Mobile / Max-Q graphics card with CUDA version 12.4 for the chaotic systems using the same settings as in \Cref{table:chaotic-system-approx}. GPU time of the double pendulum and Hénon-Heiles experiment was \( 7313.1 \) and \( 7788.9 \) seconds respectively. Using single precision has resulted in \( \approx 2.5 \) times faster training without losing a lot of accuracy. However, training HNNs using our approach was still more than 20 times faster with CPUs. 

\Cref{table:approx-params} and \Cref{table:chaotic-system-approx-params} list model parameters, and \Cref{table:domain-params} list domain parameters used in the numerical experiments in this
work. ELM bias distribution specifies the distribution used for sampling the hidden layer biases for the
ELM method. For all the experiments we use the \texttt{Uniform} distribution with \(\min\) and \(\max\)
set to the minimum and maximum domain range. Note that we do not fix this distribution to have a
more fair comparison of the weight sampling used in ELM compared to other methods, as we have noticed
a strong decline in the accuracy of ELM if the bias is sampled from a fixed range and the target domain
is scaled drastically. For the SWIM methods, we make sure that different pairs of points are selected for
different neurons for the weight construction to avoid any duplicates by re-sampling if a duplicate
is detected and repeat this until we get unique pairs, i.e., unique parameters.

\begin{table}[!htb]
  \caption{Model parameters in the single pendulum (\Cref{fig:single-pendulum} and \Cref{table:single-pendulum-lotka-volterra-experiments-summary}), single pendulum with frequency (\Cref{fig:single-pendulum-freq-errors}), and Lotka-Volterra (\Cref{fig:lotka-volterra-errors} and \Cref{table:single-pendulum-lotka-volterra-experiments-summary})
  experiments are listed. ELM bias distribution is set using the minimum and maximum
  domain boundaries. Therefore, for Lotka-Volterra experiments, it is set as \texttt{Uniform(}\(\{-2,-5,0\}, \{2,5,8\}\)\texttt{)}
  depending on the domain. The last four columns are the parameters of the traditionally trained HNN, including the total number of gradient steps in training, learning rate and weight decay in the Adam optimizer, and batch size. For Lotka-Volterra experiments, the total number of gradient steps is set to either \(15000\) or \(30000\), depending on the domain size.}\label{table:approx-params}
  \centering
  \begin{tabular}{l|lll}
    \toprule
    Parameter               &   Single pendulum                               &   Single pendulum with \(f\)                        &   Lotka-Volterra                              \\
    \midrule
    Number of layers        &   1                                             &   1                                                 &   1                                           \\
    Network width           &   See~\cref{fig:single-pendulum}                &   1500                                              &   See~\cref{fig:lotka-volterra-errors}  \\
    Activation              &   \(\tanh\)                                     &   \(\tanh\)                                         &   \(\tanh\)                                   \\
    \(L^2\) regularization  &   \(10^{-13}\)                                  &   \(10^{-13}\)                                      &   \(10^{-13}\)                                \\
    ELM bias distribution   &   \verb+Uniform(+\(-\pi,\pi\)\verb+)+           &   \verb+Uniform(+\(-\pi,\pi\)\verb+)+               &   See table caption                       \\
    HNN \#gradient-steps         &   \(15000\)               &   -                     &  See table caption \\
    HNN learning rate           &   \(5 \cdot 10^{-4}\) &   -                     &    \(5 \cdot 10^{-4}\) \\
    HNN weight decay            &   \(10^{-13}\)        &   -                     &     \(10^{-13}\) \\  
    HNN batch size              &   \(2048\)            &   -                     &     \(2048\) \\
    \bottomrule
  \end{tabular}
\end{table}

\begin{table}[!htb]
  \caption{Model parameters in the double pendulum and Hénon-Heiles experiments
  (see~\Cref{table:chaotic-system-approx}) are listed.}\label{table:chaotic-system-approx-params}
  \centering
  \begin{tabular}{l|ll}
    \toprule
    Parameter               &   Double pendulum                               &   Hénon-Heiles                        \\
    \midrule
    Number of layers        &   1                                             &   1                                   \\
    Network width           &   5000                                         &   5000                                \\
    Activation              &   \(\tanh\)                                     &   \(\tanh\)                           \\
    \(L^2\) regularization  &   \(10^{-13}\)                                  &   \(10^{-13}\)                        \\
    ELM bias distribution   &   \verb+Uniform(+\(-\pi,\pi\)\verb+)+           &   \verb+Uniform(+\(-5,5\)\verb+)+     \\
    HNN \#gradient-steps    &   \(180000\)                                        & \(180000\) \\
    HNN learning rate       &   \(10^{-4}\)                                    &  \(10^{-4}\) \\
    HNN weight decay        &   \(10^{-13}\)                                  &     \(10^{-13}\) \\
    HNN batch size          &   \(2048\)                                        &   \(2048\) 
    \\
    \bottomrule
  \end{tabular}
\end{table}

\begin{table}[!htb]
  \caption{Domain parameters of the target Hamiltonians in the experiments are listed.}\label{table:domain-params}
  \centering
  \begin{tabular}{lllll}
    \toprule
    System name                     & Hamiltonian                                                                                             & Domain                              & Train set size  & Test set size \\
    \midrule
    Single pendulum                 & \cref{eq:single-pendulum}                                                                       & \([-2\pi, 2\pi]\times[-1,1]\)       & 10000           & 10000         \\
    Single pendulum                 & \cref{eq:single-pendulum}                                                                       & \([-2\pi, 2\pi]\times[-6,6]\)       & 10000           & 10000         \\
    Single pendulum                 & \cref{eq:single-pendulum-with-frequency}                                                        & \([-\pi, \pi]\times[-0.5,0.5]\)     & See~\cref{fig:single-pendulum-freq-errors} & 10000  \\
    Lotka-Volterra                  & \cref{eq:lotka-volterra-zero-centered}                                                          & \([-2,2]\times[-2,2]\)              & 10000           & 10000         \\
    Lotka-Volterra                  & \cref{eq:lotka-volterra-zero-centered}                                                          & \([-5,5]\times[-5,5]\)              & 10000           & 10000         \\
    Lotka-Volterra                  & \cref{eq:lotka-volterra-five-centered}                                                          & \([0,8]\times[0,8]\)                & 10000           & 10000         \\
    Double pendulum                 & \cref{eq:double-pendulum}                                                                       & \([-\pi, \pi]^2\times[-1, 1]^2\)    & 20000           & 20000         \\
    Hénon-Heiles                    & \cref{eq:henon-heiles}                                                                          & \([-5,5]^2\times[-5,5]^2\)          & 20000           & 20000         \\
    \bottomrule
  \end{tabular}
\end{table}

\begin{figure}[!htb]
  \begin{center}
    \includegraphics[width=1\linewidth]{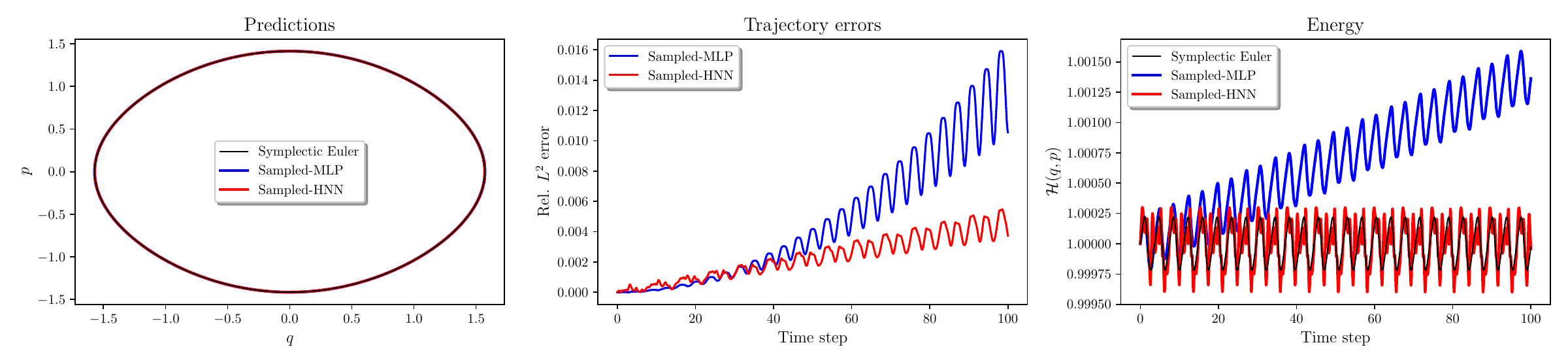}
  \end{center}
  \caption{Sampled-HNN outputs the Hamiltonian value, while Sampled-MLP directly approximates the time derivatives
  of the inputs \(\dot{q}\) and \(\dot{p}\). Single pendulum is approximated with the same domain and model
  parameters used in the larger domain single pendulum experiments (domain \([-2\pi, 2\pi]\times[-6,6]\)) and
  both Sampled-MLP and Sampled-HNN are sampled according to the ELM method. The trained models and the true Hamiltonian are
  integrated using symplectic Euler with time step size \(5\cdot10^{-4}\) from 
 the initial coordinates
  \(x_0=(\nicefrac{\pi}{2}, 0)\).}\label{fig:sampled-mlp-hnn-energy-conservation}
\end{figure}

\paragraph{Conservation of the Hamiltonian value:}
\citet{greydanus-2019-HNN} have demonstrated energy conservation of traditional
HNNs compared to plain MLPs (which directly output system dynamics \(\dot{q}\) and \(\dot{p}\)) along
trajectory predictions. We demonstrate this property using sampled networks too in
\Cref{fig:sampled-mlp-hnn-energy-conservation}.

\paragraph{Learning from trajectory data with error correction:}\label{section:appendix:numerical:error-correction}
Learning from trajectory data consisting of generalized ``position'' \( q \) and ``momentum'' \( p \) coordinates only
has been studied in many recent work. \citet{zhu-2020-HNN-with-midpointrule} theoretically analyzes
the training of traditional HNNs using finite differences with symplectic and non-symplectic integration schemes.
Networks trained with integrator schemes learn the network targets rather than the true Hamiltonian due
to discretization errors. Their work aligns with the numerical results presented by \citet{chen-2019-SRNNs},
where the trained networks account for the numerical errors. An adapted loss was proposed by \citet{zhu-2020-HNN-with-midpointrule},
which is then adapted for Gaussian Processes by \citet{offen-2022-GP-inverse-modified-hamiltonian}.
In later work, \citet{david-2023} adapted a post-training correction method to account for the discretization
errors for traditional HNNs trained with symplectic integrators to recover the true Hamiltonian function
of the target system. In the rest of this section, we incorporate this correction scheme into \Cref{eq:fully-linear-system} and demonstrate that it can also
be applied to sampled HNNs that we introduce in this paper.

\begin{figure}[!htb]
  \begin{center}
    \includegraphics[width=0.5\linewidth]{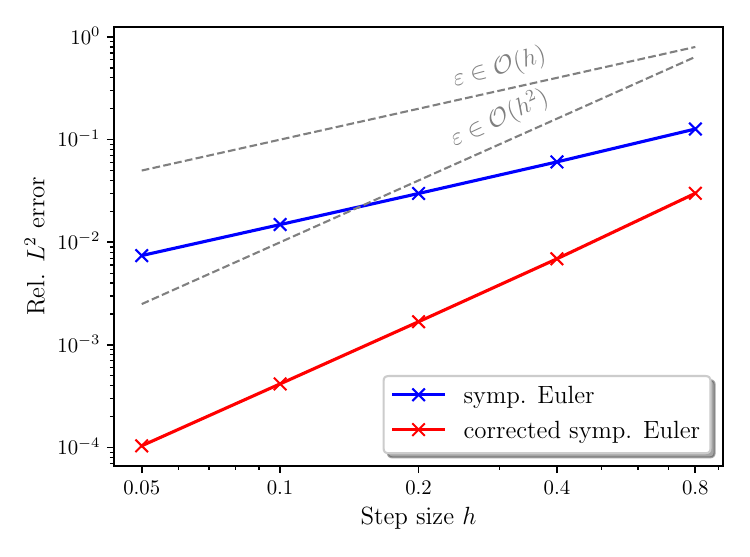}
  \end{center}
  \caption{Single pendulum Hamiltonian approximation errors are displayed with and without error correction
  together with reference lines \( \epsilon = h \) and \( \epsilon = h^2 \).
  Symp. Euler stands for the A-SWIM method using finite differences and symplectic Euler integration
  scheme as described in \Cref{eq:fully-linear-system-with-symplectic-euler-integration}.}\label{fig:error-correction}
\end{figure}

Given a trajectory dataset \( \mathcal{D} = \{x_i, \varphi_h(x_i)\}_{i=1}^{K} \) where \( \varphi_h \)
is the exact flow of the target Hamiltonian system with time step size \( \Delta t = h > 0 \).
Note that any trajectory of arbitrary length can be written to have the form of \(\mathcal{D}\) if the points
in the trajectory are considered as pairs. Similar to \citep{offen-2022-GP-inverse-modified-hamiltonian},
we adapt the linear system \Cref{eq:fully-linear-system}, which we aim to solve, to use the finite differences
on the right-hand side, and the symplectic Euler integration scheme on the left-hand side:
\begin{equation}\label{eq:fully-linear-system-with-symplectic-euler-integration}
    \begin{bmatrix}
      \nabla \Phi^{(L)}(\varphi_h(q_1), p_1) & 0 \\
      \nabla \Phi^{(L)}(\varphi_h(q_2), p_2) & 0 \\
      \vdots & \vdots \\
      \nabla \Phi^{(L)}(\varphi_h(q_K), p_K) & 0 \\
      \Phi^{(L)}(x_0) & 1
    \end{bmatrix}
  \cdot
    \begin{bmatrix}
      W_{L+1}\tran \\
      b_{L+1}
    \end{bmatrix}
  \overset{!}{=}
    \begin{bmatrix}
      J^{-1} \cdot (\nicefrac{(\varphi_h(x_1) - x_1)}{h}) \\
      J^{-1} \cdot (\nicefrac{(\varphi_h(x_2) - x_2)}{h}) \\
      \vdots \\
      J^{-1} \cdot (\nicefrac{(\varphi_h(x_K) - x_K)}{h}) \\
      \mathcal{H}(x_0)
    \end{bmatrix}.
\end{equation}
\citet{david-2023} used the well-established error analysis by \citet{hairer-2006-geometric-numerical-integration-also-symplectic}
(Chapter 9, Example 3.4) for the symplectic Euler method
\begin{equation*}
  \widehat{\mathcal{H}} = \mathcal{H} - \frac{h}{2}{\nabla_q \mathcal{H}}\tran \nabla_p \mathcal{H} + \mathcal{O}(h^2),
\end{equation*}
and corrected the discretization error of a trained traditional HNN up to an order as
\begin{equation*}
  \mathcal{H} \approx \widehat{\mathcal{H}} + \frac{h}{2}{\nabla_q \widehat{\mathcal{H}}}\tran \nabla_p \widehat{\mathcal{H}}.
\end{equation*}
We use the same correction scheme using analytical solutions for a sampled HNN \( \Phi \) as
\begin{equation*}
  \mathcal{H} \approx \Phi + \frac{h}{2}{\nabla_q \Phi}\tran \nabla_p \Phi,
\end{equation*}
and demonstrate the error correction for the single pendulum system on
domain \([-\pi, \pi]\times[-1,1]\) with train set size \( 2500 \) with their next states after time step \(h\),
test set size \(2500\), \(L^{2}\) regularization \(10^{-13}\) and network
width \( 200 \) using the A-SWIM method for the sampling in \Cref{fig:error-correction}. The true flow of the target system is simulated
using the explicit Runge-Kutta method of order 5(4) implementation from
\verb+scipy.integrate.solve_ivp+ \citep{scipy} with time step size \(10^{-4}\).

\end{document}